\documentclass[lettersize,journal]{IEEEtran}
\usepackage{algorithm}
\usepackage{algorithmic}

\usepackage{booktabs}
\usepackage{subcaption}
\usepackage{algorithmic}
\usepackage{algorithm}
\usepackage{cite}
\usepackage{diagbox}
\usepackage{thmtools,thm-restate}
\usepackage{graphicx}
\usepackage{multirow}
\usepackage{subfiles}
\usepackage{stmaryrd}
\usepackage{enumitem}
\usepackage{color}
\usepackage{diagbox}
\usepackage{multirow}
\usepackage{url}

\newcommand{\pos}[1]{[#1]_+}

\newcommand{\iter}[1]{^{(#1)}}

\newcommand{\rhos}{\rho^\src}
\newcommand{\rhot}{\rho^\tar}
\newcommand{\DB}{\mathbf{D}}

\newcommand{\RB}{\mathbf{R}}

\newcommand{\TB}{\mathbf{T}}
\newcommand{\UB}{\mathbf{U}}
\newcommand{\VB}{\mathbf{V}}

\newcommand{\WB}{\mathbf{W}}
\newcommand{\XB}{\mathbf{X}}

\newcommand{\onem}{{\mathbf{1}^m}}
\newcommand{\onen}{{\mathbf{1}^n}}
\newcommand{\aB}{\mathbf{a}}
\newcommand{\bB}{\mathbf{b}}

\newcommand{\qB}{\mathbf{q}}

\newcommand{\xB}{\mathbf{x}}
\newcommand{\yB}{\mathbf{y}}

\newcommand{\RBB}{\mathbb{R}}

\newcommand{\AM}{\mathcal{A}}
\newcommand{\BM}{\mathcal{B}}

\newcommand{\IM}{\mathcal{I}}
\newcommand{\JM}{\mathcal{J}}

\newcommand{\NM}{\mathcal{N}}
\newcommand{\OM}{\mathcal{O}}

\newcommand{\SM}{\mathcal{S}}
\newcommand{\TM}{\mathcal{T}}

\newcommand{\XM}{\mathcal{X}}

\newcommand{\epsilonB}{\mbox{\boldmath$\epsilon$\unboldmath}}
\newcommand{\gammaB}{\mbox{\boldmath$\gamma$\unboldmath}}

\newcommand{\sigmaB}{\mbox{\boldmath$\sigma$\unboldmath}}

\newcommand{\thetaB}{\mbox{\boldmath$\theta$\unboldmath}}

\newcommand{\src}{{\operatorname{s}}}
\newcommand{\tar}{{\operatorname{t}}}

\newcommand{\proj}{{\operatorname{proj}}}

\newcommand{\datasets}{datasets}
\newcommand{\elementwise}{element-wise}

\newcommand{\nonconvex}{non-convex}
\newcommand{\nonconvexity}{non-convexity}
\newcommand{\nonempty}{non-empty}

\newcommand{\nonuniform}{non-uniform}
\newcommand{\nonzero}{non-zero}

\newcommand{\realworld}{real-world}
\newcommand{\stepsize}{step-size}

\newcommand{\subproblem}{subproblem}
\newcommand{\subproblems}{subproblems}
\newcommand{\subregion}{subregion}
\newcommand{\subregions}{subregions}

\newcommand{\Realworld}{Real-world}

\usepackage{amsmath}
\usepackage{amssymb}
\usepackage{mathtools}
\usepackage{amsthm}
\theoremstyle{plain}
\newtheorem{theorem}{Theorem}[section]

\theoremstyle{definition}
\newtheorem{definition}[theorem]{Definition}

\newtheorem{example}[theorem]{Example}
\theoremstyle{remark}

\newcommand{\name}{SCOTM}
\renewcommand{\paragraph}[1]{\textbf{#1}}

\title{Many-to-Many Matching via Sparsity Controlled Optimal Transport}

\author{Weijie~Liu,~Han~Bao,~Makoto~Yamada,~Zenan~Huang,~Nenggan~Zheng,~\IEEEmembership{Senior Member,~IEEE,}~Hui~Qian
	\thanks{Manuscript received April 19, 2021; revised August 16, 2021. (\emph{Corresponding author: Nenggan Zheng.})}
	\thanks{This work is supported by National Key Research and Development Program of China under Grant 2020AAA0107400, Major Program of the National Natural Science Foundation of China (T2293723), and National Natural Science Foundation of China (Grant No: 62206248).}
	\thanks{Weijie Liu is with the Qiushi Academy for Advanced Studies, Zhejiang
		University, Hangzhou, Zhejiang 310007, China, and also with the College of
		Computer Science and Technology, Zhejiang University, Hangzhou, Zhejiang
		310007, China (e-mail: westonhunter.zju.edu.cn).\\
		\indent Han Bao is with Kyoto University, Yoshidahonmachi, Sakyo Ward, Kyoto, 606-8501, Japan (e-mail: bao@i.kyoto-u.ac.jp).\\
		\indent Makoto Yamada is with Okinawa Institute of Science and Technology, Kunigami-gun, Okinawa 904-0497, Japan, and also with RIKEN Center for Advanced Intelligence Project, Chuo-ku, Tokyo 103-0027, Japan (e-mail: makoto.yamada@oist.jp).\\
		\indent Zenan Huang is with the College of
		Computer Science and Technology, Zhejiang University, Hangzhou, Zhejiang
		310007, China (e-mail: lccurious@zju.edu.cn).\\		
		\indent Nenggan Zheng is with the Qiushi Academy for Advanced Studies, Zhejiang	University, Hangzhou, Zhejiang 310007, China, and also with the Collaborative Innovation Center for Artificial Intelligence by MOE and Zhejiang Provincial Government (ZJU) and the Zhejiang Laboratory, Hangzhou,	Zhejiang 311121, China (e-mail: zng@cs.zju.edu.cn).\\
		\indent Hui Qian is with the College of
		Computer Science and Technology, Zhejiang University, Hangzhou, Zhejiang
		310007, China, and also with State Key Lab of CAD\&CG, Zhejiang University, Hangzhou, Zhejiang
		310007, China (e-mail: qianhui@zju.edu.cn).}}%

\begin{document}

\maketitle

\begin{abstract}
	Many-to-many matching seeks to match multiple points in one set and multiple points in another set, which is a basis for a wide range of data mining problems.
	It can be naturally recast in the framework of Optimal Transport (OT).
	However, existing OT methods either lack the ability to accomplish many-to-many matching or necessitate careful tuning of a regularization parameter to achieve satisfactory results.
	This paper proposes a novel many-to-many matching method to explicitly encode many-to-many constraints while preventing the degeneration into one-to-one matching.
	The proposed method consists of the following two components.
	The first component is the matching budget constraints on each row and column of a transport plan, which specify how many points can be matched to a point at most.
	The second component is the deformed $q$-entropy regularization, which encourages a point to meet the matching budget maximally.
	While the deformed $q$-entropy was initially proposed to sparsify a transport plan, we employ it to avoid the degeneration into one-to-one matching.
	We optimize the objective via a penalty algorithm, which is efficient and theoretically guaranteed to converge.
	Experimental results on various tasks demonstrate that the proposed method achieves good performance by gleaning meaningful many-to-many matchings.
\end{abstract}


\section{Introduction}
\begin{figure}
	\includegraphics[width=\linewidth]{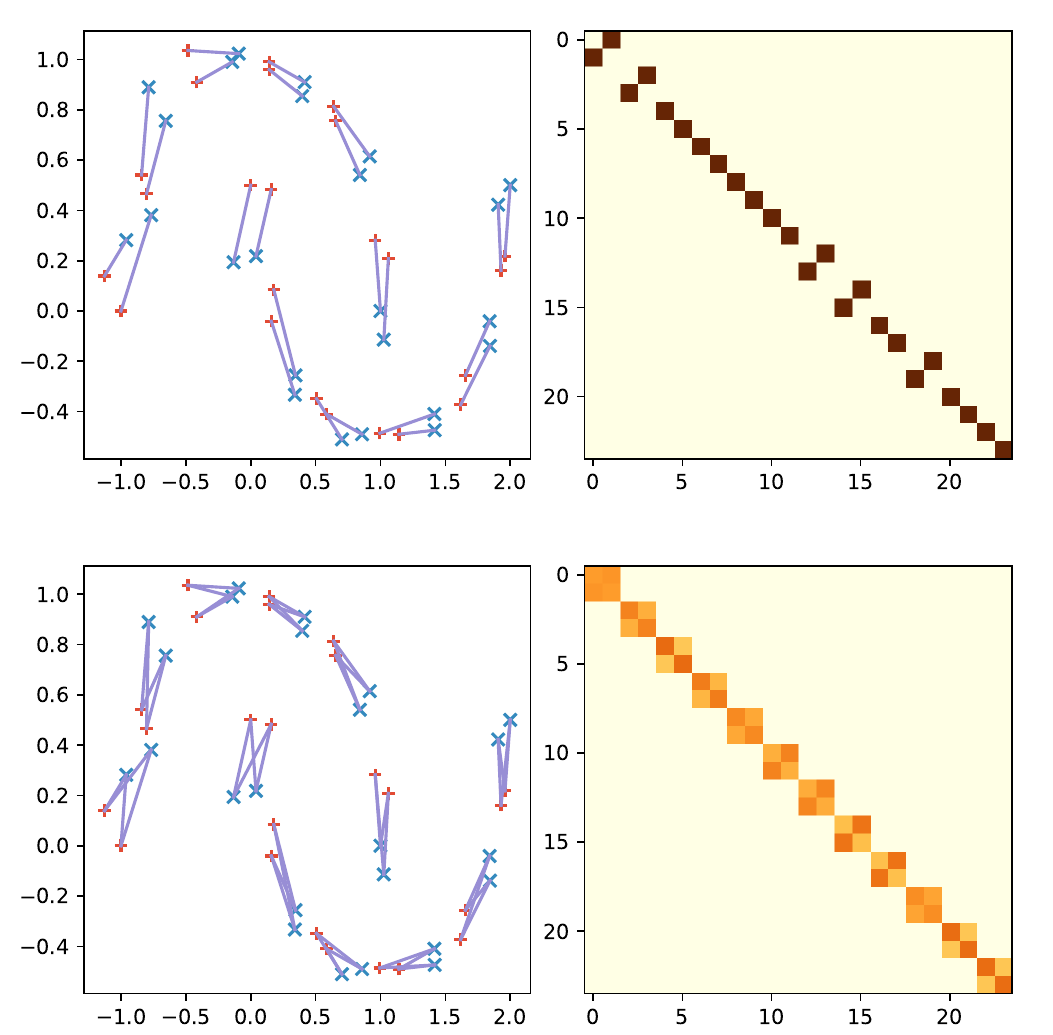}
	\caption{The matching results (left) and the transport plans (right) between two uniform discrete measures obtained by the network simplex solver for OT (top) and our method (bottom), respectively.
		The red +'s and the blue $\times$'s denote the support points of the two measures, respectively.
		The darker the color is, the larger the value of the corresponding entry in the transport plan is.
		Since the support points do not have strict one-to-one correspondence, many-to-many matching is more appropriate.}\label{fig:intro}
\end{figure}

\IEEEPARstart{M}{atching} two related sets is a fundamental problem in data mining \cite{li2014semantic,xu2023syntactic,xu2023group}.
\Realworld{} applications may require matching each point in one set to multiple points in another set, that is, many-to-many matching \cite{colannino2007efficient,rajabi2012limited,rajabi2019faster}.
Such applications include information retrieval \cite{toussaint2004comparison}, point cloud registration \cite{sun2018sparse}, associating proteins across species \cite{kalecky2018primalign}, and recommendation systems \cite{tan2022partial}.
In these applications, many-to-many matching shall be considered because no strict one-to-one correspondence exists in the data (refer to Figure~\ref{fig:intro} for an illustration), or users desire more recommended items \cite{tan2022partial}.
Mathematically speaking, many-to-many matching is formulated as finding a matching matrix by optimizing the overall utility while satisfying matching budget constraints, that is, the number of \nonzero{} entries in each row and each column are upper-bounded.
Such a problem can be naturally rephrased in the powerful and elegant framework of Optimal Transport (OT) \cite{villani2008optimal}.


Dating back to \cite{monge1781memoire}, OT compares two probability distributions by finding a transport plan between them that minimizes the overall cost \cite{kantorovich1960mathematical,villani2008optimal}.
In the past decade, OT has been employed to address a broad class of problems, including ranking \cite{yu2019wassrank,cuturi2019differentiable}, graph mining \cite{xu2019gromov,maretic2019got}, and cross-domain retrieval \cite{huang2023improving,hu2023cross}, among others.
The majority of research in this area focuses on calculating the Wasserstein distance \cite{villani2009wasserstein} or identifying a transport plan that establishes correspondence between two datasets \cite{kantorovich1960mathematical}.




While it is possible to establish many-to-many matching by selecting relatively large entries in a transport plan, the lack of a principled formulation eventually leads to brittle solutions.
A group of works attempts to simply model many-to-many matching as OT by considering points with \nonzero{} entries in a transport plan as matched pairs \cite{keselman2003many,demirci2006object,demirci2009many,demirci2011efficient}.
This post-process often results in a matching matrix with very few points being matched because solutions to the vanilla OT are usually too sparse \cite{brualdi2006combinatorial,2018computational}.
The entropy regularization has been commonly used to prevent a transport plan from being overly sparse and enhance numerical stability \cite{cuturi2013sinkhorn}, which results in a fuzzy transport plan and renders it difficult to extract an interpretable matching \cite{blondel2018smooth,xie2020fast,bao2022sparse,liu2023sparsity}.
Therefore, it is crucial to maintain a transport plan that satisfies the matching budget while avoiding the degeneration into one-to-one matching by controlling the sparsity.




To effectively and efficiently conduct many-to-many matching, this paper proposes a new method, Sparsity Controlled Optimal Transport for Many-to-many matching (\name{}), which consists of two components.
First, to enforce the matching budget, we impose the sparsity constraints on both the rows and columns of a transport plan.
Second, to prevent a transport plan from being overly sparse, we regularize the transport plan with the deformed $q$-entropy that was initially proposed to sparsify a transport plan \cite{bao2022sparse}.
\name{} encourages as many points to be matched as possible under the matching budget by adjusting the value of $q$.
This novel application of the deformed $q$-entropy is of independent interest.
We then propose a penalty algorithm to decompose the intricate feasible domain into the intersection of four sets, each of which admits a closed-form update of the projection step.
Our contributions can be summarized as follows:
\begin{enumerate}
	\item We introduce a practical OT-based many-to-many matching formulation, which imposes the row/column sparsity constraints and promotes matches using the $q$-entropy regularization.
	\item We develop an efficient penalty algorithm to handle the complex feasible domain and solve the underlying optimization problem of \name.
	\item We prove that the sequence of the transport plan output by the penalty algorithm admits a limit point, which satisfies the first-order optimality condition.
\end{enumerate}
Empirical results in a variety of tasks demonstrate that \name{} outperforms state-of-the-art methods.
Specifically, synthetic experiments illustrate that \name{} can control the transport plan sparsity more flexibly than existing OT variants.
Experiments in student course allocation, matching protein interaction networks, and object recognition show that \name{} performs well in a wide range of \realworld{} applications.

The rest of the paper is organized as follows.
In the next section, a comprehensive review of the background is given.
Section~\ref{sec:methodology} delineates the proposed many-to-many matching method.
Experimental results on various tasks are reported in Section~\ref{sec:experiment}.
The appendix provides detailed theoretical proofs.


\section{Preliminary and Related Work}\label{sec:preliminary}

\subsection{Notation}
For $x\in\RBB$, let $\pos{x}=x$ if $x>0$ and $0$ otherwise.
We use bold lowercase symbols (e.g. $\xB$), bold uppercase letters (e.g. $\XB$), uppercase calligraphic fonts (e.g. $\XM$), and Greek letters (e.g. $\alpha$) to denote vectors, matrices, spaces (sets), and measures, respectively.
The $d$-dimensional all-ones vector is denoted by $1^d \in \mathbb{R}^d$.
The probability simplex with $d$ bins is denoted by $\Delta^d$, namely  $\Delta^d=\big\{\aB\in\RBB^d_{+}\mid\sum_{i=1}^{d}a_i=1\big\}$.
The expression $\xB\ge c$ (resp. $\XB\ge c$) indicates that each entry of the vector $\xB$ (resp. the matrix $\XB$) is greater than or equal to the scalar $c$.
In the matrix $\XB$, $\XB[i,:]$ represents the $i$\textsuperscript{th} row, and $\XB[:,j]$ denotes the $j$\textsuperscript{th} column.
Given a matrix $\XB$, expressions $\|\XB\|_F$, $\|\XB\|_0$, and $\|\XB\|_\infty$ denote its Frobenius norm, number of \nonzero{} elements, and maximum absolute value (i.e., $\|\XB\|_\infty = \max_{i,j}|X_{ij}|$), respectively.
The cardinality of set $\XM$ is denoted by $|\XM|$.
The bracketed notation $\llbracket n\rrbracket$ is the shorthand for the integer set $\{1,2,\dots,n\}$.
A discrete measure $\alpha$ is denoted by $\alpha=\sum_{i=1}^{m}a_i\delta_{\xB_i}$, where $\delta_{\xB}$ is the Dirac measure at position $\xB$, i.e., a unit of mass infinitely concentrated at $\xB$.
When $\alpha$ is a probability measure, $\aB=[a_i]$ is in the simplex $\aB\in\Delta^m$.

\subsection{Many-to-many Matching}

Over the past few decades, many-to-many matching has received a lot of attention in numerous fields including combinatorial optimization \cite{zhu2016solving,das2019gradient}, operations research \cite{hamidouche2014many,biro2021complexity,xepapadeas2022serial}, and economics \cite{echenique2004theory,chen2019pareto,pkeski2022tractable}.
Given two disjoint sets $\SM=\{s_1,s_2,\dots,s_m\}$ and $\TM=\{t_1,t_2,\dots,t_n\}$, many-to-many matching involves finding a binary matching matrix $\TB=[T_{ii'}]\in\{0,1\}^{m\times n}\cap\Omega_{\rho^\src,\rho^\tar}$ between $\SM$ and $\TM$ satisfying prescribed budget constraints denoted by $\Omega_{\rho^\src,\rho^\tar}$, i.e., $T_{ii'}=1$ if $s_i$ is matched to $t_{i'}$ and $T_{ii'}=0$ otherwise, and $\TB\in\Omega_{\rho^\src,\rho^\tar}$ indicates that $\TB$ has at most $\rho^\src$ and $\rho^\tar$ \nonzero{} entries in each row and column, respectively:
\begin{equation*}
	\begin{aligned}
		\Omega_{\rho^\src,\rho^\tar}=\Big\{\TB\in\RBB_+^{m\times n}\,\big|\,&\big\|\TB[i,:]\big\|_0\le\rho^\src\text{ for all } i,\\
		&\text{ and }\big\|\TB[:, i']\big\|_0\le\rho^\tar\text{ for all } i'\Big\}.
	\end{aligned}
\end{equation*}
Typical examples of many-to-many matching include the following scenarios:
\begin{enumerate}
	\item Curriculum management system: Each student (from set $\SM$) has a limit $\rhos$ on the number of courses he or she can attend, while each course (from set $\TM$) has a limit $\rhot$ on the number of students that it can admit \cite{cechlarova2018pareto,utture2019student}.
	This scenario represents a combinatorial optimization challenge within an educational context, where the goal is to optimally allocate courses to students based on their preferences and course capacities.
	\item Matching Protein Interaction Networks (PINs): Proteins and their interactions are at the core of almost every biological process.
	In a PIN, nodes represent proteins and edges correspond to interactions between proteins.
	By matching PINs across different species, it is possible to establish correspondences between proteins, thereby performing protein function predictions and transferring protein knowledge from well-studied species like mice to other species \cite{singh2008global}.
	\item Task assignment: Project management systems are ubiquitous in various organizations where tasks and agents are represented as two distinct sets. One task can be assigned to multiple agents, and conversely, an agent may undertake multiple tasks \cite{zhu2016solving,das2019gradient}.
	A many-to-many matching method can be used to efficiently assign tasks to agents, considering constraints like the capability of agents and the requirements of each task.
	\item Video caching: In a bandwidth-constrained network, base stations select videos to cache according to their local popularity to reduce end-users' experienced delay \cite{hamidouche2014many}.
	Each video can be cached by multiple base stations and each base station can cache multiple videos, which allows the video caching system to better balance the load on the network and improve the overall user experience.
\end{enumerate}
Existing non-OT many-to-many matching methods primarily fall into the following two categories.

\paragraph{Multi-objective optimization.}
Some methods formulate many-to-many matching as multi-objective optimization problems \cite{cechlarova2014pareto,cechlarova2016pareto,cechlarova2018pareto,chen2019pareto,bando2021substitutes}.
By modeling each point as an autonomous agent that has a preference ranking over points from the other set, these methods match each point to as high-ranking points as possible while achieving Pareto optimality and stability. 
Nevertheless, these methods usually have high computational complexities because each iteration requires finding a Pareto improvement, which involves $\OM(m^2n^2)$ time \cite{cechlarova2014pareto,chen2019pareto}.


\paragraph{Single-objective optimization.}
Several studies formulated many-to-many matching based on single objective optimization by encoding the preferences with a preference matrix $\RB=[R_{ii'}]$, as follows \cite{colannino2007efficient,rajabi2012limited,rajabi2013two}:
\begin{equation}\label{eq:many-to-many}
		\max_{\TB\in\{0,1\}^{m\times n}\cap\Omega_{\rho^\src,\rho^\tar}}\sum_{i=1}^{m}\sum_{i'=1}^{n} R_{ii'}T_{ii'}.
\end{equation}
To solve the problem (\ref{eq:many-to-many}), it has been common to use the Hungarian method \cite{zhu2009mm,rajabi2013two,zhu2016solving} or solve its continuous relaxation with the interior point method \cite{das2019gradient}.
These approaches have computational complexity $\OM\big(\max\{m,n\}^3\max\{\rho^\src,\rho^\tar\}^3\big)$, which is more efficient than multi-objective optimization methods when $m$ and $n$ are of the same order, and both $\rhos$ and $\rhot$ are relatively small.
However, when $\rhos$ or $\rhot$ is large, they can also be time-consuming.
In addition, a naive solution to the problem (\ref{eq:many-to-many}) may be \emph{excessively sparse} such that each point matches to only a very small number of points.
Even if such a solution is feasible for (\ref{eq:many-to-many}), we desire to obtain a solution that matches as many points as possible under the budget constraints.

Another limitation of these non-OT methods is that they treat all points in $\SM$ (resp. $\TM$) with equal importance. 
In some situations, we may need to prioritize matching certain points.
For example, in task assignment, some tasks may be more urgent or important, necessitating prioritized resource allocation.
Though this might be achievable in single-objective optimization methods by modifying the preference matrix in an ad-hoc manner, such modifications typically demand careful control.
This limitation can be naturally and effectively addressed by our OT-based many-to-many matching method that is derived in Section~\ref{sec:methodology}.
We first review OT in the following subsection.

\subsection{Optimal Transport and Its Variants}

\paragraph{Optimal Transport.}
Given discrete probability measures $\alpha=\sum_{i=1}^m a_i\delta_{\xB_i}$ and $\beta=\sum_{i'=1}^n b_{i'}\delta_{\yB_{i'}}$, OT is formulated by the following optimization problem:
\begin{equation}\label{eq:ot_projective}
	\min_{\TB\in\Pi(\aB,\bB)}\Big\{\ell(\TB):=\sum_{i=1}^{m}\sum_{i'=1}^{n} c(\xB_i,\yB_{i'})T_{ii'}\Big\},
\end{equation}
where $c(\cdot,\cdot)$ is a cost function and the feasible domain of \emph{transport plan} $\TB=[T_{ii'}]$ is given by the set
\begin{equation*}
	\Pi(\aB,\bB)=\{\TB\in\RBB_+^{m\times n}\mid\TB\onen=\aB,\TB^\top\onem=\bB\}.
\end{equation*}
The optimal objective value of (\ref{eq:ot_projective}) is often referred to as the OT distance \cite{villani2008optimal,2018computational}.
The problem (\ref{eq:ot_projective}) is a linear program and can be solved by off-the-shelf linear programming methods \cite{tarjan1997dynamic,ge2019interior} which, however, incur cubic computational complexities and are not scalable to large \datasets{}.

\paragraph{Entropy regularized OT.}
To enhance the scalability, \cite{cuturi2013sinkhorn} proposed to regularize a transport plan with the negative Shannon entropy as follows:
\begin{equation*}
	\min_{\TB\in\Pi(\aB,\bB)}\ell(\TB)-\gamma H(\TB),
\end{equation*}
where $\gamma > 0$ is the regularization weight and $H(\TB)=-\sum_{i=1}^{m}\sum_{i'=1}^{n}(T_{ii'}\log T_{ii'}-T_{ii'})$ is the negative Shannon entropy.
\cite{cuturi2013sinkhorn} showed that the regularized problem can be solved by the Sinkhorn algorithm, whose computational complexity is quadratic.
However, the resulting transport plan is strictly positive and completely dense, which is undesirable in many applications where a transport plan is of interest \cite{blondel2018smooth,xie2020fast}.

\paragraph{Quadratic regularized OT.}
\cite{blondel2018smooth} instead considered the quadratic regularization as follows:
\begin{equation*}
	\min_{\TB\in\Pi(\aB,\bB)}\ell(\TB)+\gamma\sum_{i'=1}^{n}\big\|\TB[:,i']\big\|_2^2.
\end{equation*}
It is shown that the quadratic regularization can induce sparsity in the resulting transport plan.


\paragraph{$q$-regularized OT.}
\cite{bao2022sparse} further introduced the \emph{deformed $q$-entropy}, which is defined as follows:
\begin{equation*}
	\resizebox{\linewidth}{!}{$
		H_q(\TB)=\begin{cases}
			&\frac{-1}{2-q}\sum_{i=1}^{m}\sum_{i'=1}^{n}\Big(\frac{T_{ii'}^{2-q}-T_{ii'}}{1-q}-T_{ii'}\Big),\text{ if }q\in[0,1),\\
			&-\sum_{i=1}^{m}\sum_{i'=1}^{n}(T_{ii'}\log T_{ii'}-T_{ii'}),\text{ if }q=1.
		\end{cases}
		$}
\end{equation*}
The deformed $q$-entropy requires $q\in[0,1]$ and recovers the negative Shannon entropy and quadratic regularization at $q=1$ and $q=0$, respectively.
As $q$ approaches $1$, the solution becomes less sparse, and it becomes completely dense if $q=1$.
Therefore, by using the deformed $q$-entropy as regularization, the sparsity of a transport plan can be controlled by varying the parameter $q$, as follows:
\begin{equation*}
	\min_{\TB\in\Pi(\aB,\bB)}\ell(\TB)-\gamma H_q(\TB).
\end{equation*}
In this work, we heavily leverage the sparsity control with the deformed $q$-entropy to obtain a transport plan that maximally satisfies the row/column budget constraints of many-to-many matching.

\paragraph{Sparsity-constrained OT.}
\cite{liu2023sparsity} applied explicit cardinality constraints on the columns of a transport plan as follows:
\begin{equation*}
	\min_{\TB\in\Pi(\aB,\bB)}\ell(\TB)+\gamma\sum_{i'=1}^{n}\big\|\TB[:,i']\big\|_2^2,\text{ s.t. }\forall i', \big\|\TB[:,i']\big\|_0\le\rho.	
\end{equation*}
Such a formulation enables direct control of the sparsity of each column in a transport plan.
However, it cannot be directly used for many-to-many matching, because ({\bf i}) there is no cardinality constraint applied in the row direction, which may result in too many \nonzero{} entries in some rows, and ({\bf ii}) the resulting transport plan is likely to be overly sparse, which leads to nearly one-to-one matching.

\subsection{Further Related Work}

\begin{figure}[t]
	\includegraphics[width=\linewidth]{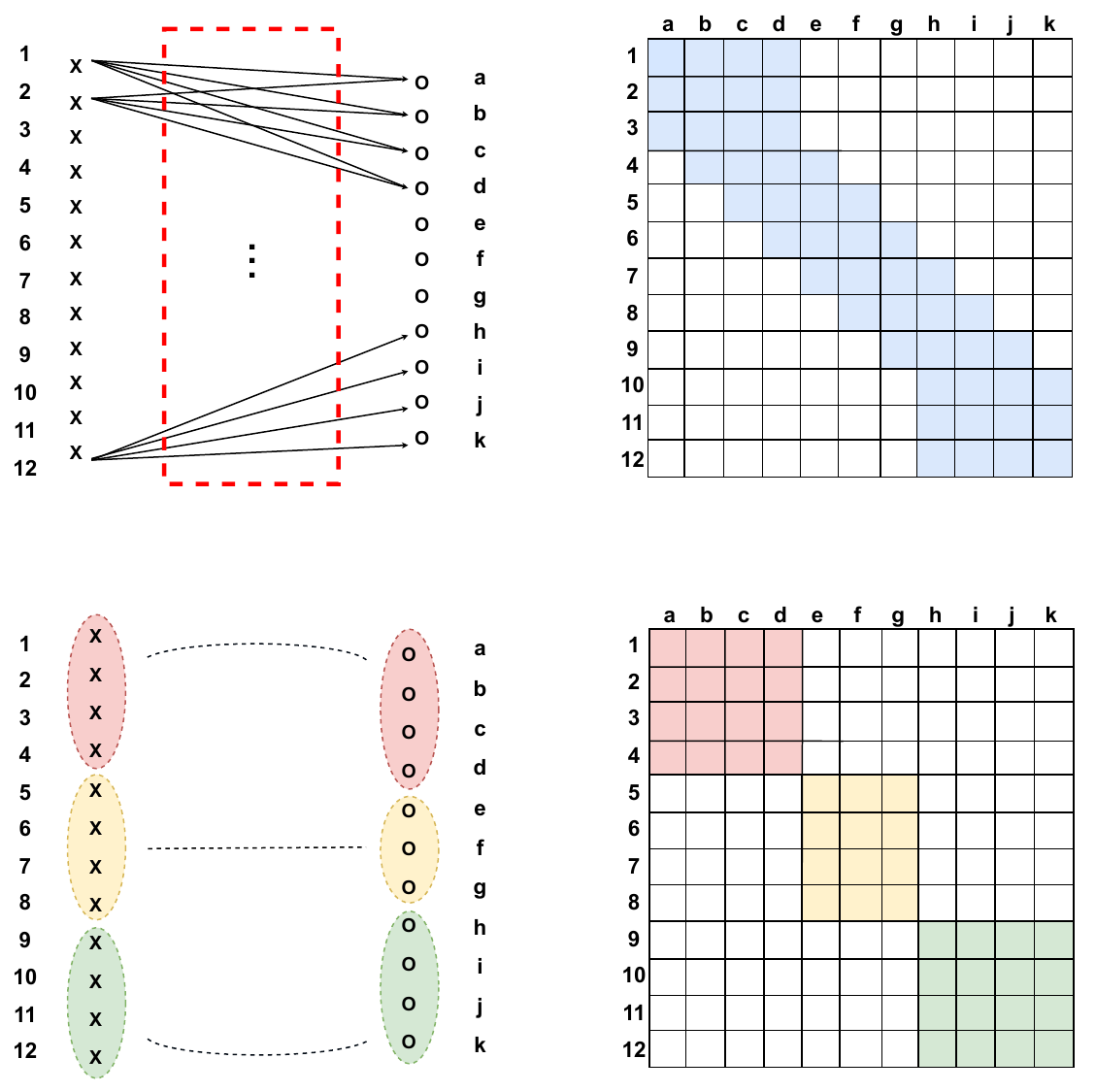}
	\caption{A comparison between many-to-many matching (top) and cluster-to-cluster matching (bottom) in the student course allocation problem.
		The sets of students and courses are denoted as $\{1,2,\dots,12\}$ and $\{a,b,\dots,k\}$, respectively.
		Each student can take up to 4 courses and each course can admit up to 6 students.
		In the cluster-to-cluster matching, the students are divided into the three clusters (groups), and all students in each cluster take exactly the same courses.}\label{fig:many_vs_cluster}
\end{figure}


\paragraph{Cluster-to-cluster matching.}
A closely related problem is cluster-to-cluster matching \cite{zaslavskiy2010many,iwata2017robust,iwata2017topic,weylandt2021sparse}, which requires simultaneous assignment of each point to a cluster and determination of the one-to-one correspondence between clusters.\footnote{Although some of these works also use the terminology ``many-to-many matching", this paper refers to this type of matching as cluster-to-cluster matching to distinguish the two notions.}
If a cluster $\IM\subset\SM$ is matched to a cluster $\IM'\subset\TM$, all points in the cluster $\IM$ are matched to the cluster $\IM'$, and the specific points matching within the clusters are not of concern.
A cluster-to-cluster matching can also be decomposed into two separate many-to-one matches, that is, matching the two sets to a virtual set \cite{zaslavskiy2010many}.
The feasible domain of cluster-to-cluster matching is a subset of the feasible domain of many-to-many matching when the cardinalities of sets $\IM$ and $\IM'$ are constrained by $|\IM|\le\rhot$ and $|\IM'|\le\rhos$ respectively, which demonstrates that many-to-many matching can accommodate more flexible matching structures.
See Figure~\ref{fig:many_vs_cluster} for an illustration in the context of a student-course allocation problem.
In addition, some clusters may contain a single point, which results in one-to-one or many-to-one matching for these points.


\paragraph{OT-based one-to-one matching.}
Many works in the OT literature involve obtaining one-to-one matching from the calculated transport plan \cite{solomon2016entropic,xu2019gromov,maretic2019got}.
To this end, they solve a linear assignment problem based on the transport plan, or simply choose the largest entry in each row (or column) of the transport plan to indicate the matching.
One-to-one matching methods are characterized by matching matrices with at most one \nonzero{} entry per row and column, which is a special case of many-to-many matching when $\rhos=1$ and $\rhot=1$.
Some works further seek to incorporate structures of probability distributions into a transport plan \cite{alvarez2018structured,lin2021making,lim2022order}.
Closely related to our work, \cite{alvarez2018structured} proposed to encourage correlations in matchings of source supports that are close, which however does not control the sparsity of a transport plan explicitly.

\section{Methodology}\label{sec:methodology}
In this section, we first present our OT-based many-to-many matching formulation.
Then, an optimization algorithm is derived.
We finally demonstrate the convergence guarantee of the proposed algorithm.
For conciseness, all detailed proofs are deferred to the appendix.

\subsection{Problem Formulation}
First, we recast many-to-many matching into the following OT problem with the additional budget constraints:
\begin{equation}\label{eq:unregularized}
	\min_{\TB}\ell(\TB),\text{ s.t. }\TB\in\Pi(\aB,\bB)\cap\Omega_{\rho^\src,\rho^\tar},
\end{equation}
which, as is implied by the following theorem, is well-defined under suitable assumptions.
\begin{restatable}{theorem}{primenonempty}\label{thm:nonempty}
	Let $\JM'(h)$ (resp. $\JM(h)$) be a subset of $\llbracket n\rrbracket$ (resp. $\llbracket m\rrbracket$) with cardinality $h$.
	When the following conditions hold:
	\begin{equation}\label{eq:nonempty1}
			\|\aB\|_\infty\le\min_{\JM'(\rhos-1)\subset\llbracket n\rrbracket}\sum_{i'\in\JM'(\rhos-1)}b_{i'},
	\end{equation}
	and
	\begin{equation}\label{eq:nonempty2}
		\|\bB\|_\infty\le\min_{\JM(\rhot-1)\subset\llbracket m\rrbracket}\sum_{i\in\JM(\rhot-1)}a_{i},
	\end{equation}
	the feasible domain $\Pi(\aB,\bB)\cap\Omega_{\rho^\src,\rho^\tar}$ is \nonempty{}.
\end{restatable}
Therefore, (\ref{eq:unregularized}) can be regarded as a well-defined continuous alternative of (\ref{eq:many-to-many}) when $\rhos$ and $\rhot$ satisfy (\ref{eq:nonempty1}) and (\ref{eq:nonempty2}), making it possible to solve many-to-many matching problems more efficiently.
By the Extreme Value Theorem \cite{rudin1976principles}, the continuous objective function over the \nonempty{} and compact feasible domain $\Pi(\aB,\bB)\cap\Omega_{\rho^\src,\rho^\tar}$ has a minimum.
Once a solver for (\ref{eq:unregularized}) outputs a transport plan, many-to-many matching can simply be indicated by its \nonzero{} entries.
Another advantage of (\ref{eq:unregularized}) over (\ref{eq:many-to-many}) is that it can set \nonuniform{} distributions $\aB$ and/or $\bB$ to prioritize some points, as is formalized in the following theorem.
\begin{restatable}{theorem}{primepriority}\label{thm:priority}
	Assume that $\aB$ and $\bB$ satisfy (\ref{eq:nonempty1}) and (\ref{eq:nonempty2}).
	When
	\begin{equation}\label{eq:priority_a_condition}
		a_{i}\ge\max_{\JM'(h)\subset\llbracket n\rrbracket}\sum_{i'\in\JM'(h)}b_{i'},
	\end{equation}
	where $h\in\llbracket\rhos-1\rrbracket$, at least $h$ points are matched to $s_i$.
	Similarly, when
	\begin{equation}\label{eq:priority_b_condition}
		b_{i'}\ge\max_{\JM(h)\subset\llbracket m\rrbracket}\sum_{i\in\JM(h)}a_{i},
	\end{equation}
	where $h\in\llbracket\rhot-1\rrbracket$, at least $h$ points are matched to $t_{i'}$.
\end{restatable}
Theorem \ref{thm:priority} implies that our formulation can ensure a flexible lower bound $h$ of the number of points matched to a prioritized point.
More specifically,  the design of $\aB$ and $\bB$ has to firstly satisfy the conditions in Theorem \ref{thm:nonempty} to ensure that the feasible domain is \nonempty{}.
Secondly, $\aB$ (resp. $\bB$) needs to satisfy Eq. (\ref{eq:priority_a_condition}) (resp. (\ref{eq:priority_b_condition})) to guarantee that at least $h$ points are matched to prioritized points in $\SM$ (resp. $\TM$), which is demonstrated in the following example.

\begin{example}
	Without loss of generality, the proportion of prioritized points in $\SM$ is denoted by $r$, and assume that at least $h$ points need to be matched to prioritized points.
	Then, $\bB$ can still be set as a uniform distribution while $\aB=[a_i]$ can be chosen as follows:
	\begin{equation}\label{eq:priority_a}
		a_i=\begin{cases}
			&\frac{h}{n}\text{, if $i$ is a prioritized point,}\\
			&\frac{1-mrh/n}{m-mr}\text{, otherwise,}
		\end{cases}
	\end{equation}
	where $h\in\llbracket \rhos-1\rrbracket$.
	It is obvious that $\sum_{i=1}^m a_i=1$.
	Under such a setting, as long as $m$, $n$, and $h$ satisfy
	\begin{equation}\label{eq:priority_verification}
		n\le mh,
	\end{equation}
	(\ref{eq:nonempty1}), (\ref{eq:nonempty2}) and (\ref{eq:priority_a_condition}) hold simultaneously.
	Note that Eq. (\ref{eq:priority_b_condition}) is not required to hold since only some points in $\SM$ are prioritized.
\end{example}




\paragraph{Deformed $q$-entropy regularization.}
One limitation of the formulation (\ref{eq:unregularized}) is that the budget constraints entail only the upper bounds of the row/column cardinalities, and thus even a feasible solution can be so sparse that it degenerates into nearly one-to-one matching.
Hence, an additional technique is needed to encourage matching as many points as possible.
We propose to adopt the deformed $q$-entropy regularization to address this issue by considering the following regularized problem:
\begin{equation}\label{eq:final_objective}
	\min_{\TB\in\Pi(\aB,\bB)\cap\Omega_{\rho^\src,\rho^\tar}}\Big\{G(\TB):=\ell(\TB)-\gamma H_q(\TB)\Big\},
\end{equation}
where a value close to $1$ should be chosen for $q$ to induce as dense a solution as possible under the budget constraints.
Our formulation achieves many-to-many matching by controlling the sparsity of transport plans, and is thus referred to as Sparsity Controlled Optimal Transport for Many-to-many matching (SCOTM).
It is worth noting that the common negative Shannon entropy is inapplicable, because the gradients of the regularized objective with respect to a transport plan are ill-defined in the feasible region of the problem (\ref{eq:final_objective}).

\paragraph{Remark.}
Problem (\ref{eq:final_objective}) involves a convex optimization objective and a \nonconvex{} feasible domain.
One may wonder whether it is possible to relax (\ref{eq:final_objective}) by discarding the \nonconvex{} $\Omega_{\rho^\src,\rho^\tar}$ and using a convex sparsity-inducing regularizer (e.g., the $\ell_1$ norm).
Unfortunately, such an approach does not grant us direct control over the number of \nonzero{} entries in each row and column. 
Moreover, adding the $\ell_1$ norm regularizer of a transport plan is ineffective, because it is equivalent to adding a constant to the objective and does not influence the solution, that is,
\begin{equation*}
	\begin{aligned}
		&\min_{\TB\in\Pi(\aB,\bB)}G(\TB)+\lambda\|\TB\|_1=\lambda+\min_{\TB\in\Pi(\aB,\bB)}G(\TB).
	\end{aligned}
\end{equation*}




\begin{algorithm}[t]
	\caption{Penalty Algorithm for the problem (\ref{eq:final_objective}).}\label{alg:penalty}
	\begin{algorithmic}[1]
		\STATE {\bf Input:} Initialized variables $\TB\iter{0}=\UB\iter{0}=\VB\iter{0}=\WB\iter{0}$ s.t. $\TB\iter{0}\in\Pi(\aB,\bB)\cap\Omega_{\rho^\src,\rho^\tar}$, scalars $\sigma\iter{0}>0$ and $\theta>1$, and a sequence $\{\epsilon\iter{k}\}_k$ s.t. $\epsilon\iter{k}\to 0$.
		\STATE {\bf Output:} The sequence $\{\TB\iter{k}\}$.
		\FOR{$k=0,1,\dots$}
		\STATE Calculate the \elementwise{} partial gradient of $J_{\sigma\iter{k}}(\cdot)$ w.r.t. $\TB\iter{k}$:
		\begin{equation*}
			\DB\iter{k}=\frac{\partial J_{\sigma\iter{k}}(\TB\iter{k},\UB\iter{k},\VB\iter{k},\WB\iter{k})}{\partial \TB},
		\end{equation*}
		and find \stepsize{} $\eta$ via the Armijo line search in the direction $-\DB\iter{k}$.
		\STATE Calculate $\TB_{\operatorname{trial}}=\proj_{\Omega^1}\big(\TB\iter{k}-\eta\DB\iter{k}\big)$.
		\IF{$J_{\sigma\iter{k}}\big(\TB_{\operatorname{trial}},\UB\iter{k},\VB\iter{k},\WB\iter{k}\big)$ $\le$ $G(\TB\iter{0})$}
		\STATE Set $\TB\iter{k,0}=\TB\iter{k}$, $\UB\iter{k,0}=\UB\iter{k}$, $\VB\iter{k,0}=\VB\iter{k}$, and $\WB\iter{k,0}=\WB\iter{k}$.
		\ELSE
		\STATE Set $\TB\iter{k,0}=\TB\iter{0}$, $\UB\iter{k,0}=\UB\iter{0}$, $\VB\iter{k,0}=\VB\iter{0}$, and $\WB\iter{k,0}=\WB\iter{0}$.
		\ENDIF
		\STATE Set $l=0$.
		\WHILE{$\|\TB\iter{k,l+1}-\TB\iter{k,l}\|_F>\epsilon\iter{k}$}
		\STATE\label{line:inner_start} Find \stepsize{} $\eta$ via the Armijo line search in the direction $-\DB\iter{k,l}$.
		\STATE Calculate $\TB\iter{k,l+1}$ as Eq. (\ref{eq:update_T}).
		\STATE Calculate $\UB\iter{k,l+1}$, $\VB\iter{k,l+1}$, and $\WB\iter{k,l+1}$ independently as Eq. (\ref{eq:update_UVW}).
		\STATE\label{line:inner_end} Update $l=l+1$.
		\ENDWHILE
		\STATE Set $\sigma\iter{k+1}=\theta\sigma\iter{k}$.
		\STATE Set $\TB\iter{k+1}=\TB\iter{k,l}$, $\UB\iter{k+1}=\UB\iter{k,l}$, $\VB\iter{k+1}=\VB\iter{k,l}$, and $\WB\iter{k+1}=\WB\iter{k,l}$.
		\ENDFOR
	\end{algorithmic}
\end{algorithm}

\subsection{Optimization}

Unlike \cite{blondel2018smooth}, \cite{bao2022sparse}, and \cite{liu2023sparsity} that solved their respective dual problems, we directly consider the primal problem and express (\ref{eq:final_objective}) by the following equivalent problem:
\begin{equation}\label{eq:decomposition}
	\resizebox{.88\linewidth}{!}{$
		\displaystyle
		\min_{\substack{\TB\in\Omega^1,\UB\in\Omega^2,\VB\in\Omega^3,\WB\in\Omega^4}} G(\TB), \text{ s.t. }\TB=\UB=\VB=\WB,
		$}
\end{equation}
where
\begin{equation*}
	\begin{aligned}
		\Omega^1&=\{\TB\in\RBB_+^{m\times n}\mid\TB\onen=\aB\},\\
		\Omega^2&=\{\UB\in\RBB_+^{m\times n}\mid\UB^\top\onem=\bB\},\\
		\Omega^3&=\big\{\VB\in\RBB_+^{m\times n}\mid\big\|\VB[i,:]\big\|_0\le\rho^\src\text{ for all } i\in\llbracket m\rrbracket\big\},\\
		\Omega^4&=\big\{\WB\in\RBB_+^{m\times n}\mid\big\|\WB[:,i']\big\|_0\le\rho^\tar\text{ for all } i'\in\llbracket n\rrbracket\big\}.
	\end{aligned}
\end{equation*}
As we shall see shortly, such a constraint decomposition allows closed-form updates of all projection steps, whereas dual optimization does not admit an analytical solution in general.
For example, we need to resort to solving a quadratic regularized OT problem that does not enjoy a closed-form solution unless $\Pi(\aB, \bB)$ is decomposed into the two constraint sets $\Omega^1$ and $\Omega^2$.

\begin{figure*}[t]
	\includegraphics[width=\linewidth]{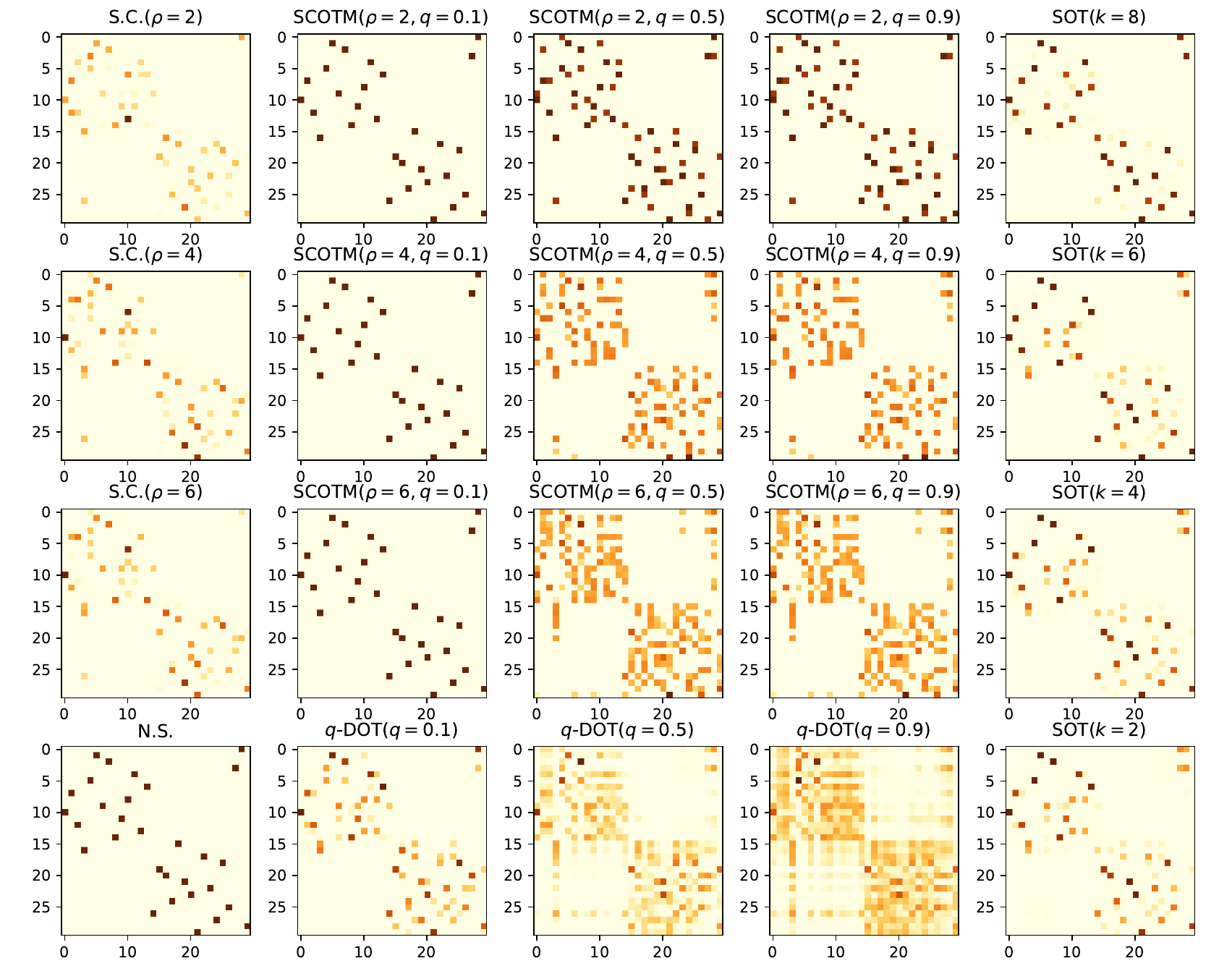}
	\caption{A comparison of the transport plans obtained by the network simplex solver (N.S.), sparsity-constrained OT (S.C.), $q$-regularized OT ($q$-DOT), structured OT (SOT), and \name.
		The darker the color is, the larger the value of the corresponding entry is in the transport plan.}\label{fig:plans}
\end{figure*}

After the reformulation (\ref{eq:decomposition}), a penalty algorithm is applied.
In particular, we consider a sequence of penalty \subproblems{}, with the $k$-th \subproblem{} given by
\begin{equation*}
	\min_{\TB\in\Omega^1,\UB\in\Omega^2,\VB\in\Omega^3,\WB\in\Omega^4}J_{\sigma\iter{k}}(\TB,\UB,\VB,\WB),
\end{equation*}
where
\begin{equation*}
	\begin{aligned}
		J_{\sigma\iter{k}}(\TB,\UB,\VB,\WB):&=G(\TB)+\frac{\sigma\iter{k}}{2}\|\TB-\UB\|^2_F\\
		&+\frac{\sigma\iter{k}}{2}\|\TB-\VB\|^2_F+\frac{\sigma\iter{k}}{2}\|\TB-\WB\|^2_F,
	\end{aligned}
\end{equation*} 
which is solved approximately by updating the four variables alternately.
Such an approach is summarized in Algorithm \ref{alg:penalty}, which involves an outer loop and an inner loop.
We use the index $k$ for the outer iteration and $l$ for the inner iteration.
Each outer iteration corresponds to a \subproblem{}, and the details of inner iterations are described as follows.


\paragraph{Updating $\TB$.}
In each iteration $l$, $\TB\iter{k,l}$ is updated via the projected gradient descent with fixed $\UB\iter{k,l}$, $\VB\iter{k,l}$, and $\WB\iter{k,l}$ as follows:
\begin{equation}\label{eq:update_T}
	\TB\iter{k,l+1}=\proj_{\Omega^1}\Big(\TB\iter{k,l}-\eta\DB\iter{k,l}\Big),
\end{equation}
where $\eta$ is the \stepsize{} and $\DB\iter{k,l}$ is the \elementwise{} partial derivative of $J_{\sigma\iter{k}}(\cdot)$ w.r.t. $\TB\iter{k,l}$:
\begin{equation*}
	\DB\iter{k,l}=\frac{\partial J_{\sigma\iter{k}}(\TB\iter{k,l},\UB\iter{k,l},\VB\iter{k,l},\WB\iter{k,l})}{\partial \TB}.
\end{equation*}
Instead of manually setting the \stepsize{}, the \stepsize{} $\eta$ is found via the Armijo line search, which is crucial to guarantee the monotone decrease of $J_{\sigma\iter{k}}(\cdot)$.
The projection operation after the gradient descent projects each row $i$ of $\TB\iter{k,l}-\eta\DB\iter{k,l}$ onto a ``scaled" simplex $\AM_i=\{a_i \xB\mid\xB\in\Delta^n\}$, which requires $\OM(mn\log n)$ time to find the closed-form solution (for further details about projecting a vector onto a simplex, see \cite{wang2013projection}).
The per-iteration complexity for updating $\TB$ is thus $\OM(mn\log n)$.

\paragraph{Updating $\UB$, $\VB$, and $\WB$.}
By the form of $J_{\sigma\iter{k}}(\cdot)$, the updates of $\UB$, $\VB$, and $\WB$ can be conducted independently:
\begin{equation}\label{eq:update_UVW}
	\min_{\UB\in\Omega^2,\VB\in\Omega^3,\WB\in\Omega^4}J_{\sigma\iter{k}}(\TB\iter{k,l+1},\UB,\VB,\WB),
\end{equation}
which essentially projects $\TB\iter{k,l+1}$ onto $\Omega^2$, $\Omega^3$, and $\Omega^4$, respectively.
$\UB\iter{k,l+1}$ is obtained by projecting each column $i'$ of $\TB\iter{k,l+1}$ onto a scaled simplex $\BM_{i'}=\{b_{i'}\xB\mid\xB\in\Delta^m\}$, which incurs cost $\OM(mn\log m)$.
Despite the \nonconvexity{} of $\Omega^3$ (resp. $\Omega^4$), $\VB\iter{k,l+1}$ (resp. $\WB\iter{k,l+1}$) can be determined by finding the top-$\rho^\src$ (resp. $\rho^\tar$) entries of each row (resp. column) of $\TB\iter{k,l+1}$, which involves per-iteration complexity $\OM(mn\log \rho^\src)$ (resp. $\OM(mn\log \rho^\tar)$) \cite{liu2023sparsity}.
Overall, the per-iteration complexity for updating the four variables is $\OM\big(mn\log\max\{m,n\}\big)$.

\paragraph{Remark.}
\name{} can also be generalized to accommodate different matching budgets for different points, mutatis mutandis.
For this variant, the optimization procedure remains valid since closed-form updates are still applicable, and the theorems have analogous versions, although the proofs will be lengthier.
For succinctness, we have focused on the setting where all points in $\SM$ (resp. $\TM$) have the same matching budget.

\paragraph{Convergence analysis.}
To present the convergence guarantee of Algorithm \ref{alg:penalty}, we define the following first-order optimality condition of (\ref{eq:final_objective}), which is adapted from \cite{lu2013sparse} and \cite{lapucci2021convergent} to accommodate the constraints in our problem.
\begin{definition}
	$\hat{\TB}$ satisfies the \emph{first-order optimality condition} of (\ref{eq:final_objective}) if $\hat{\TB}\in\Omega_{\rho^\src,\rho^\tar}\text{ and }\sum_{ii'}\frac{\partial G(\hat{\TB})}{\partial T_{ii'}}(T'_{ii'}-\hat{T}_{ii'})\ge0\text{ for all } \TB'\in\Pi(\aB,\bB)\cap\Omega_{\rho^\src,\rho^\tar}$.
\end{definition}


Based on the first-order optimality condition, the following theorem gives a theoretical guarantee of Algorithm \ref{alg:penalty}.
\begin{restatable}{theorem}{primetheorem}\label{thm:convergence}
	Algorithm \ref{alg:penalty} satisfies:
	\begin{enumerate}
		\item For each outer iteration $k\ge0$, Algorithm \ref{alg:penalty} generates $(\TB\iter{k+1},\UB\iter{k+1},\VB\iter{k+1},\WB\iter{k+1})$ in a finite number of inner iterations;
		\item The sequence $\{\TB\iter{k},\UB\iter{k},\VB\iter{k},\WB\iter{k}\}_k$ admits a limit point $(\hat{\TB},\hat{\UB},\hat{\VB},\hat{\WB})$;
		\item Every limit point satisfies that $\hat{\TB}=\hat{\UB}=\hat{\VB}=\hat{\WB}$ and $\hat{\TB}$ is a first-order optimal solution.
	\end{enumerate}
\end{restatable}
Theorem \ref{thm:convergence} demonstrates that Algorithm \ref{alg:penalty} is well-defined, and the sequence generated by it admits a limit point that satisfies the first-order optimality condition.
In practice, Algorithm \ref{alg:penalty} can be terminated when $\TB\iter{k}$, $\UB\iter{k}$, $\VB\iter{k}$, and $\WB\iter{k}$ are sufficiently close to each other.

\section{Experiments}\label{sec:experiment}

\begin{table*}[t]
	\centering
	\caption{The percentage of \nonzero{} entries in the transport plans obtained by the network simplex solver of OT (N.S.), sparsity-constrained OT (S.C.), $q$-regularized OT ($q$-DOT), structured OT (SOT), and \name.}\label{tab:sparsity}
	\setlength{\tabcolsep}{1.8mm}{
		\begin{tabular}{c | cccc | ccc | ccc | ccc | ccc | ccc}
			\toprule
			\multirow{3}{*}[0pt]{\bf N.S.} & \multicolumn{4}{c|}{\bf SOT}                                                                           & \multicolumn{3}{c|}{\bf $q$-DOT}                                                    & \multicolumn{3}{c|}{\bf S.C.}                                                          & \multicolumn{9}{c}{\bf \name{}}                                                                   \\
			\cline{2-20} 
			& \multirow{2}{*}{$k$=$8$} & \multirow{2}{*}{$k$=$6$} & \multirow{2}{*}{$k$=$4$} & \multirow{2}{*}{$k$=$2$} & \multirow{2}{*}{$q$=$.1$} & \multirow{2}{*}{$q$=$.5$} & \multirow{2}{*}{$q$=$.9$} & \multirow{2}{*}{$\rho$=$2$} & \multirow{2}{*}{$\rho$=$4$} & \multirow{2}{*}{$\rho$=$6$} & \multicolumn{3}{c|}{$\rho$=$2$} & \multicolumn{3}{c|}{$\rho$=$4$} & \multicolumn{3}{c}{$\rho$=$6$} \\
			&                        &                        &                        &                        &                          &                          &                          &                           &                           &                           & $q$=$.1$  & $q$=$.5$ & $q$=$.9$ & $q$=$.1$  & $q$=$.5$ & $q$=$.9$ & $q$=$.1$  & $q$=$.5$ & $q$=$.9$ \\
			\midrule
			$3.3$                 & $57.1$                   & $59.4$                   & $73.4$                   & $82.9$                   & $8.8$                     & $33.2$                     & $98.9$                     & $5.9$                       & $6.3$                       & $6.3$                       & $4.8$      & $6.7$     & $6.7$     & $7.4$     & $13.2$    & $13.3$    & $7.5$     & $19.4$    & $20.0$  \\
			\bottomrule 
		\end{tabular}
	}
	
\end{table*}

\begin{table*}[t]
	\centering
	\caption{The performance of many-to-many matching methods in student course allocation with various $\rhos$.}\label{tab:student_course_allocation}
	\begin{tabular}{l | rrr | rrr | rrr}
		\toprule
		& \multicolumn{3}{c|}{$\rhos=10$}                                                        & \multicolumn{3}{c|}{$\rhos=20$}                                                        & \multicolumn{3}{c}{$\rhos=40$}                                                        \\
		& {\bf \name} & {\bf KMB} & {\bf LBF} & {\bf \name} & {\bf KMB} & {\bf LBF} & {\bf \name} & {\bf KMB} & {\bf LBF} \\
		\midrule
		{\bf top-}$\mathbf{1}$ (\%)  & $\mathbf{22.1}$     & $20.8$     & $11.8$     & $\mathbf{32.7}$    & $\mathbf{32.7}$     & $12.6$      & $\mathbf{49.3}$    & $\mathbf{49.3}$      & $17.7$     \\
		{\bf top-}$\mathbf{5}$ (\%)  & $\mathbf{62.3}$     & $52.0$     & $40.3$     & $\mathbf{81.1}$    & $75.7$     & $39.7$      & $\mathbf{93.9}$    & $92.8$      & $66.0$     \\
		{\bf Runtime (sec.)} & $\mathbf{12.2}$     & $247.6$    & $400.5$    & $\mathbf{12.8}$    & $943.4$    & $1452.8$    & $\mathbf{12.9}$   & $6259.8$    & $7599.6$               \\      
		\bottomrule
	\end{tabular}
\end{table*}

\begin{table*}[t!]
	\centering
	\caption{The performance of \name{} in student course allocation with various values of $q$.}\label{tab:student_course_allocation_q}
	\begin{tabular}{r | r r r r r r r}
		\toprule
		$\qB$     & $\mathbf{0}$ & $\mathbf{0.1}$ & $\mathbf{0.5}$ & $\mathbf{0.8}$  & $\mathbf{0.9}$  & $\mathbf{0.99}$ & $\mathbf{0.999}$ \\
		\midrule
		{\bf top-}$\mathbf{1}$ (\%) & $\mathbf{32.7}$ & $\mathbf{32.7}$ & $\mathbf{32.7}$ & $\mathbf{32.7}$ & $\mathbf{32.7}$ & $\mathbf{32.7}$ & $\mathbf{32.7}$  \\
		{\bf top-}$\mathbf{5}$ (\%) & $65.4$ & $66.2$ & $76.8$ & $76.8$ & $\mathbf{81.1}$ & $\mathbf{81.1}$ & $\mathbf{81.1}$ \\
		\bottomrule
	\end{tabular}
\end{table*}

\begin{table*}[t!]
	\centering
	\caption{The performance of \name{} in student course allocation with various values of $\gamma$.}\label{tab:student_course_allocation_gamma}
	\begin{tabular}{r | r r r r r r r r r r}
		\toprule
		$\gammaB$ & $\mathbf{0.01}$ & $\mathbf{0.02}$ & $\mathbf{0.05}$ & $\mathbf{0.1}$  & $\mathbf{0.2}$  & $\mathbf{0.5}$  & $\mathbf{1}$  & $\mathbf{2}$ & $\mathbf{5}$  & $\mathbf{10}$   \\
		\midrule
		{\bf top-}$\mathbf{1}$ (\%)  & $\mathbf{32.7}$ & $\mathbf{32.7}$ & $\mathbf{32.7}$ & $\mathbf{32.7}$ & $\mathbf{32.7}$ & $\mathbf{32.7}$ & $\mathbf{32.7}$ & $30.5$ & $30.0$ & $29.8$ \\
		{\bf top-}$\mathbf{5}$ (\%)   & $73.7$ & $76.8$ & $\mathbf{81.1}$ & $\mathbf{81.1}$ & $\mathbf{81.1}$ & $\mathbf{81.1}$ & $\mathbf{81.1}$ & $75.7$ & $74.1$ & $73.7$ \\
		\bottomrule
	\end{tabular}
\end{table*}

\begin{table*}[t!]
	\centering
	\caption{Runtime (sec.) of \name{} in student course allocation with varying $\sigma\iter{0}$, $\theta$, and $\epsilon\iter{k}$.}\label{tab:runtime}
	\begin{tabular}{rr | rrr | rrr | rrr}
		\toprule
		\multicolumn{2}{c|}{$\sigmaB\mathbf{\iter{0}}$}                     & \multicolumn{3}{c|}{$\mathbf{1}$} & \multicolumn{3}{c|}{$\mathbf{10}$} & \multicolumn{3}{c}{$\mathbf{100}$} \\
		\multicolumn{2}{c|}{$\thetaB$}                     & $\mathbf{1.2}$    & $\mathbf{2.0}$   & $\mathbf{2.5}$  & $\mathbf{1.2}$    & $\mathbf{2.0}$   & $\mathbf{2.5}$   & $\mathbf{1.2}$     & $\mathbf{2.0}$   & $\mathbf{2.5}$   \\
		\midrule
		\multirow{4}{*}{\bf Bases of $\epsilonB\mathbf{\iter{k}}$} & $\mathbf{0.999}$ & $27.1$   & $14.1$  & $\mathbf{11.4}$ & $25.4$   & $12.4$  & $\mathbf{10.3}$  & $8.6$     & $4.5$   & $\mathbf{4.1}$   \\
		& $\mathbf{0.99}$  & $28.7$   & $14.1$  & $11.5$ & $25.6$   & $12.8$  & $10.4$  & $8.8$     & $4.6$   & $4.1$   \\
		& $\mathbf{0.9}$   & $78.3$   & $15.4$  & $11.5$ & $44.6$   & $12.8$  & $10.6$  & $9.9$     & $5.0$   & $4.2$   \\
		& $\mathbf{0.8}$   & $395.4$  & $31.2$  & $13.3$ & $103.3$  & $15.0$  & $10.7$  & $37.0$    & $5.2$   & $4.6$  \\
		\bottomrule
	\end{tabular}
\end{table*}

\begin{table}[t]
	\caption{PINs of C.elegans (CE), D.melanogaster (DM), H.sapiens (HS), M.musculus (MM), and S.cerevisiae (SC).}\label{tab:gm_dataset}	
	\centering
	\begin{tabular}{l r r r r r}
		\toprule
		{\bf PINs} & {\bf CE}     & {\bf DM}     & {\bf HS}     & {\bf MM}     & {\bf SC}     \\
		\midrule
		\# nodes     & 19,756 & 14,098 & 22,369 & 24,855 & 6,659  \\
		\# edges & 4,884  & 25,054 & 55,168 & 592    & 82,932 \\
		\bottomrule
	\end{tabular}
	
\end{table}

\begin{table*}[t]
	\centering
	\caption{Recall, Precision, and F$1$ (in percent) of various methods in PIN matching experiment.}\label{tab:gm_results}
	\begin{tabular}{l | r r r | r r r | r r r}
		\toprule
		\multirow{2}{*}{\bf Methods} & \multicolumn{3}{c|}{\bf CE$\to$SC} & \multicolumn{3}{c|}{\bf CE$\to$DM} & \multicolumn{3}{c}{\bf HS$\to$MM} \\
		& Recall     & Precision    & F$1$      & Recall     & Precision    & F$1$      & Recall     & Precision    & F$1$      \\
		\midrule
		{\bf PrimAlign}                & 1.1      & {\bf 62.7}     & 2.2     & 5.8      & {\bf 86.7}     & 10.9    & 0.9      & {\bf 42.6}     & 1.7     \\
		{\bf NetCoffee2}               & 3.9      & 5.2      & 4.4     & 2.4      & 3.2      & 2.7     & 1.2      & 2.7      & 1.6     \\
		{\bf N.S.}                     & 0.3      & 43.9     & 0.5     & 0.3      & 45.6     & 0.5     & 0.1      & 24.5     & 0.2     \\
		{\bf S.C.}                     & 0.1      & 10.4     & 0.2     & 0.1      & 43.3     & 0.1     & 0.1      & 11.8     & 0.1     \\
		{\bf $q$-DOT}                  & {\bf 100.0}      & 5.3      & 10.1    & {\bf 100.0}      & 3.2      & 6.2     & {\bf 100.0}      & 2.7      & 5.3     \\
		{\bf \name }    & 45.1     & 31.1     & {\bf 36.8}    & 47.3     & 32.2     & {\bf 38.3}    & 23.7     & 27.2     & {\bf 25.3}   \\
		\bottomrule
	\end{tabular}
\end{table*}

\begin{table*}[t]
	\centering
	\caption{Runtime (in sec.) of various methods in PIN matching experiment.}\label{tab:gm_runtime}
	\begin{tabular}{l | r r r r r r}
		\toprule
		{\bf Methods} & {\bf PrimAlign} & {\bf NetCoffee2} &{\bf N.S.}   & {\bf S.C.}    & {\bf $q$-DOT} & {\bf \name }   \\
		\midrule
		{\bf CE$\to$SC}    & 1762.3    & 3346.0    & 6.9    & 1934.3  & 287.7   & 1079.2  \\
		{\bf CE$\to$DM}    & 14807.8   & 13288.1    & 33.9   & 8744.0  & 3211.2  & 2878.0  \\
		{\bf HS$\to$MM}    & 107916.0  & 64736.9    & 1351.2 & 20347.9 & 11314.3 & 17703.2\\
		\bottomrule
	\end{tabular}
\end{table*}

\begin{table}[t]
	\centering
	\caption{Recall, Precision, and F$1$ (in percent) of \name{} with various $\rho$'s in PIN matching experiment.}\label{tab:gm_rhos}
	\begin{tabular}{l | r r r r r r}
		\toprule
		$\mathbf{\rho}$ & {\bf 32}  & {\bf 64}  & {\bf 128}  & {\bf 256}  & {\bf 512}  & {\bf 1024} \\
		\midrule
		Recall              & 0.9 & 2.6 & 7.7  & 25.9 & 45.1 & {\bf 69.5} \\
		Precision           & 9.7 & 7.1 & {\bf 42.5} & 35.9 & 31.1 & 24.0 \\
		F$1$                & 1.6 & 3.8 & 13.3 & 30.1 & {\bf 36.8} & 35.7\\
		\bottomrule
	\end{tabular}
\end{table}

\begin{table}[t]
	\centering
	\caption{Recognition rate (in percent) of  as a function of increasing perturbation in feature matching-based object recognition experiment. Methods include Network Simplex solver of OT (N.S.), sparsity-constrained OT (S.C.), $q$-regularized OT ($q$-DOT), and \name{}.}\label{tab:recognition_results}
	\begin{tabular}{lllll}
		\toprule
		{\bf Perturbation} & $0\%$    & $10\%$   & $20\%$   & $30\%$   \\
		\midrule
		{\bf N.S.}           & $85.2$ & $81.1$ & $73.7$ & $67.7$ \\
		{\bf S.C.}         & $87.4$ & $81.5$ & $77.8$ & $70.3$ \\
		{\bf $q$-DOT}      & $44.4$ & $38.3$ & $33.3$ & $30.9$ \\
		{\bf \name{}}        & $\mathbf{87.7}$ & $\mathbf{85.2}$ & $\mathbf{81.5}$ & $\mathbf{79.0}$\\
		\bottomrule
	\end{tabular}
\end{table}

\begin{table}[t]
	\centering
	\caption{Recognition rate (in percent) of \name{} with varying values of $\rho$ and $\gamma$ when the perturbation is 30\%.}\label{tab:recognition_sensitivity}
	\begin{tabular}{l | lllll}
		\toprule
		\diagbox{$\mathbf{\rho}$}{$\mathbf{\gamma}$} & $\mathbf{0.01}$ & $\mathbf{0.02}$ & $\mathbf{0.05}$ & $\mathbf{0.1}$  & $\mathbf{0.2}$  \\
		\midrule
		$\mathbf{4}$  & $79.0$ & $79.0$ & $77.9$ & $77.9$ & $76.6$ \\
		$\mathbf{8}$  & $79.0$ & $79.0$ & $79.0$ & $77.9$ & $76.6$ \\
		$\mathbf{16}$ & $79.0$ & $79.0$ & $79.0$ & $77.6$ & $75.3$ \\
		\bottomrule
	\end{tabular}
\end{table}

\begin{table*}[t!]
	\centering
	\caption{Empirical results on prioritized task assignment.}\label{tab:prioritized_task_assignment}
	\begin{tabular}{l | l | r r r r r}
		\toprule
		$\mathbf{n}$                  & {\bf metrics}        & {\bf KMB}     & {\bf LBF}     & {\bf \name{} ($\mathbf{h=8}$)} & {\bf \name{} ($\mathbf{h=7}$)} & {\bf \name{} ($\mathbf{h=6}$)} \\
		\midrule
		\multirow{3}{*}{$\mathbf{32}$}  & {\bf PPPM (\%)}      & $8.8$     & $11.3$    & $\mathbf{23.7}$         & $20.6$         & $15.7$         \\
		& {\bf PSMBPP (\%)}    & $51.9$    & $66.7$    & $\mathbf{100.0}$        & $\mathbf{100.0}$        & $88.9$         \\
		& {\bf Runtime (sec.)} & $1.5$     & $22.1$    & $1.0$          & $0.9$          & $0.8$          \\
		\midrule
		\multirow{3}{*}{$\mathbf{128}$} & {\bf PPPM (\%)}      & $9.5$     & $8.4$     & $\mathbf{24.8}$         & $21.2$         & $16.9$         \\
		& {\bf PSMBPP (\%)}    & $52.1$    & $46.2$    & $\mathbf{99.2}$         & $96.6$         & $92.3$         \\
		& {\bf Runtime (sec.)} & $131.4$   & $350.1$   & $1.9$          & $1.6$          & $1.8$          \\
		\midrule
		\multirow{3}{*}{$\mathbf{512}$} & {\bf PPPM (\%)}      & $9.8$     & $9.1$     & $\mathbf{29.1}$         & $17.0$         & $16.5$         \\
		& {\bf PSMBPP (\%)}    & $54.5$    & $50.8$    & $\mathbf{100.0}$        & $94.8$         & $91.9$         \\
		& {\bf Runtime (sec.)} & $29355.4$ & $11919.9$ & $27.6$         & $26.1$         & $25.2$        \\
		\bottomrule
	\end{tabular}
\end{table*}



In this section, we empirically verify the effectiveness of \name{} on various tasks.
In all tasks, the outer loops are terminated when
\begin{equation*}
	\resizebox{\linewidth}{!}{$
		\displaystyle
	\sqrt{\|\TB\iter{k}-\UB\iter{k}\|^2_F+\|\TB\iter{k}-\VB\iter{k}\|^2_F+\|\TB\iter{k}-\WB\iter{k}\|^2_F}\le10^{-4}.
	$}
\end{equation*}
Throughout this section, we choose $\sigma\iter{0}=10$, $\theta=2.0$, and $\epsilon\iter{k}=0.99^k\times10^{-4}$ for hyperparameters in Algorithm \ref{alg:penalty}, unless specified otherwise.
The experiments are conducted on a Ubuntu 18.04 server with two 28-core 2.40 GHz Intel\textsuperscript{\tiny\textregistered} Xeon\textsuperscript{\tiny\textregistered} E5-2680 v4 CPUs and 378 GB of RAM.
The source code is written in Python 3.6 and is included in the supplementary material.
The package dependencies are listed in the requirements.txt file in the code submission.

\subsection{Matching Synthetic Point Clouds}

\paragraph{Experimental setup.}
In the first set of experiments, we seek to understand the characteristics of transport plans obtained by \name.
We synthesize two point clouds in $\RBB^2$ from the two-Gaussian mixture $0.5\NM\big([\begin{smallmatrix}
	0\\
	0
\end{smallmatrix}],[\begin{smallmatrix}
	1 & 0\\
	0 & 1
\end{smallmatrix}]\big)+0.5\NM\big([\begin{smallmatrix}
	2\\
	2
\end{smallmatrix}],[\begin{smallmatrix}
	1 & 0\\
	0 & 1
\end{smallmatrix}]\big)$.
Both point clouds contain $30$ points.
The cost function $c(\cdot,\cdot)$ is fixed to the squared Euclidean distance.
Both the source and target distributions are uniform.

\paragraph{Baselines.}
The baselines include the network simplex solver for OT \cite{tarjan1997dynamic}, sparsity-constrained OT \cite{liu2023sparsity}, $q$-regularized OT \cite{bao2022sparse}, and structured OT \cite{alvarez2018structured} that divides support points into \subregions{} and encourages correlations of the matching within the same \subregion{}.
For structured OT, we used $k$-means clustering to obtain \subregions{}.
We used the same value $\rho$ for both $\rho^\src$ and $\rho^\tar$, and tested different values of $\rho$ for \name.
The regularization weight, $\gamma=0.1$, was shared by sparsity-constrained OT, $q$-regularized OT, and \name.

\paragraph{Sparsity.}
The transport plans obtained by different methods are visualized in Figure \ref{fig:plans} and the corresponding \nonzero{} entry percentages are reported in Table \ref{tab:sparsity}.
The transport plan obtained by the network simplex solver is maximally sparse.
For sparsity-constrained OT and $q$-regularized OT, the transport plan becomes denser with $\rho$ and $q$ increasing, respectively.
The transport plan of structured OT becomes sparser as $k$ increases because fewer transport plan entries are correlated, however, it is hard to directly control the sparsity.
\name{} obtains sparser transport plans than $q$-regularized OT with the same $q$ especially when $q$ is relatively large.
Owing to the adoption of the deformed $q$-entropy regularization, \name{} outputs denser transport plans than sparsity-constrained OT when $\rho=4$ or $\rho=6$.
Figure \ref{fig:plans} and Table \ref{tab:sparsity} showcase that \name{} can control the sparsity of transport plans directly and flexibly.
Specifically, while \name{} uses parameters $q$, $\rhos$, and $\rhot$ to control the sparsity of transport plans, practitioners can simply set $q$ close to $1$, and adjust matching budget constraints $\rhos$ and $\rhot$ to achieve the required sparsity and thus the desired number of matched pairs.
Such a property is essential to extract interpretable many-to-many matching and cannot be achieved by naive OT or its existing variants.
This unique property makes \name{} user-friendly.

\subsection{Student Course Allocation}

\paragraph{Experimental setup.}
We proceed to consider the student course allocation problem using the \realworld{} Harvard course allocation dataset \cite{budish2012multi}.
This dataset, obtained through a survey, comprises the preference rankings of 456 students for 96 courses.
The preference score for each student's $k$\textsuperscript{th} favorite course is assigned as $\frac{1}{k}$ and the corresponding matching cost is $1-\frac{1}{k}$.
We compared \name{} with $q = 0.9$ and $\gamma=0.1$, against state-of-the-art many-to-many matching algorithms, including KMB \cite{zhu2016solving} and LBF \cite{das2019gradient}.
The performance for different values of $q$ and $\gamma$ is discussed in the appendix.
Assuming that each student can select at most $\rhot=5$ courses, we evaluate the percentage of $k$\textsuperscript{th} favorite courses being selected (referred to as top-$k$) and the runtime of these methods for various $\rhos$ values.

\paragraph{Results in student course allocation.}
As demonstrated in Table \ref{tab:student_course_allocation}, \name{} exhibits superior computational efficiency compared to KMB and LBF.
For instance, when each course permits at most 40 students, \name's runtime is $99.79\%$ and $99.83\%$ less than that of KMB and LBF, respectively.
On both the top-$1$ and top-$5$ metrics, \name{} achieves performance that is similar to or better than KMB and LBF.
To further understand the performance of \name{} on the student course allocation task, we conduct sensitivity analyses of $q$, $\gamma$, $\sigma\iter{0}$, $\theta$, and $\epsilon\iter{k}$ while fixing $\rhos=20$ and $\rhot=5$.

\paragraph{Sensitivity analysis of $q$.}
As is demonstrated in Table \ref{tab:student_course_allocation_q}, when $q\ge0.9$, \name{} is not sensitive to the choice of $q$.
When $q$ is too small, the transport plan is likely to be too sparse and the top-$5$ metric tends to decrease, which justifies our usage of deformed $q$-entropy to promote matches.

\paragraph{Influence of $\gamma$.}
We further study the impact of the regularization weight $\gamma$ by fixing $q=0.9$ and  varying values of $\gamma$.
As is shown in Table \ref{tab:student_course_allocation_gamma}, \name{} is insensitive to $\gamma$ and achieves the best performance when $\gamma\in[0.05,1]$.
When $\gamma$ gets too small, top-$5$ decreases but top-$1$ does not, which is due to the fact that the transport plan becomes sparse.
On the other hand, if $\gamma$ is too large, both top-$1$ and top-$5$ deteriorate as the matching costs contribute insufficiently.

\paragraph{Influence of $\sigma\iter{0}$, $\theta$, and $\epsilon\iter{k}$.}
To understand the roles of $\sigma\iter{0}$, $\theta$, and $\epsilon\iter{k}$, We run \name{} with $\sigma\iter{0}\in\{1,10,100\}$, $\theta\in\{1.2,2.0,2.5\}$, and $\epsilon\iter{k}\in 10^{-4}\times\{0.999^k,0.99^k,0.9^k,0.8^k\}$.
Table \ref{tab:runtime} demonstrates that larger $\sigma\iter{0}$ and $\theta$ and larger bases of $\epsilon\iter{k}$ tend to reduce time.
However, the matching quality is not sensitive to them since all cases yield top-$5$ values between $80.3\%$ and $81.1\%$ and top-$1$ remains a constant $32.7\%$.

\subsection{Matching Protein Interaction Networks}

\paragraph{Experimental setup.}
Five extensively studied species are considered \cite{sahraeian2013smetana,alkan2014beams,hu2019novel}, including \emph{C.elegans} (worm), \emph{D.melanogaster} (fly), \emph{H.sapiens} (human), \emph{M.musculus} (mouse), and \emph{S.cerevisiae} (yeast).\footnote{These datasets can be downloaded from \url{http://webprs.khas.edu.tr/~cesim/BEAMS.tar.gz}.}
Table \ref{tab:gm_dataset} summarizes the statistics of these datasets.
We evaluate the performance of various methods using the hierarchical Gene Ontology (GO) categorization, where proteins are annotated with appropriate GO categories organized as a directed acyclic graph.
As is the common practice in the literature \cite{singh2008global,alkan2014beams}, we restrict the protein annotations to level five of the GO directed acyclic graph.
A protein pair that shares at least one annotation is considered as a true match.
The sequence similarity of proteins is used to set the corresponding matching cost.
We compare \name{} against other OT-based methods and state-of-the-art many-to-many graph matching methods including PrimAlign \cite{kalecky2018primalign} and NetCoffee2 \cite{hu2019novel}.
For $q$-regularized OT and \name{}, we chose $q=0.9$.
We set regularization weight $\gamma=0.1$ and used the same value $\rho$ for both $\rhos$ and $\rhot$.

\paragraph{Results in matching protein interaction networks.}
Table \ref{tab:gm_results} reports the results when $\rho=512$.
\name{} achieves a good balance between recall and precision, and outperforms baselines by a large margin in terms of F1.
Both the network simplex solver and sparsity-constrained OT exhibit low recall values because their transport plans are overly sparse.
In contrast, the transport plans of $q$-regularized OT are overly dense.
Table \ref{tab:gm_results} summarizes the runtime of various methods.
\name{} has similar runtime to sparsity-constrained OT and $q$-regularized OT, and has higher computational efficiency than PrimAlign and NetCoffee2.
We further compared the performance of \name{} with various $\rho$'s, which is reported in Table \ref{tab:gm_rhos}.
Increasing $\rho$ can increase recall since \name{} outputs more matched pairs.
On the other hand, the precision may decrease when $\rho$ is too large.

\subsection{Object Recognition}

\paragraph{Experimental setup.}
In this subsection, we conduct a shape retrieval experiment on the benchmark COIL-20 dataset that includes 20 objects with 72 views each.
Following \cite{demirci2006object}, we begin by removing 36 (of the 72) representative views of each object (every other view), and use these removed views as queries to the remaining view database (the other 36 views for each of the 20 objects).
Each object's silhouette is then represented by a discrete skeleton, i.e., a set of points supported on two-dimensional Euclidean spaces.
We then perform PCA-based alignment for obtained points and compute the discrepancies between each ``query" view and each of the remaining database views according to the optimized objectives of optimal transport and its variants.
For any given query view, we predict its object using the object of the database view that has the smallest discrepancy.
We use the same $\rho=8$ value for both $\rhos$ and $\rhot$.
The regularization weight $\gamma=0.01$ is shared by all OT variants.
We set $q=0.9$ for both $q$-regularized OT and \name.
To also perform perturbation studies, we randomly remove 10\%, 20\%, and 30\% of points.

\paragraph{Results in object recognition.}
The detailed experimental results are reported in Table \ref{tab:recognition_results}.
Our method achieves the best performance and is more robust to the perturbation than baselines.
Furthermore, the sensitivity analysis in Table \ref{tab:recognition_sensitivity} demonstrates that our method is insensitive to the choice of $\rho$ and $\gamma$.

\subsection{Prioritized Task Assignment}


\paragraph{Experimental setup.}
Next, we verify the effectiveness of \name{} for prioritized matching on a group of synthetic \datasets{}.
In this group of \datasets{}, the cardinalities of $\SM$ and $\TM$ are equal, i.e., $m=n$.
We compare \name{} with state-of-the-art many-to-many matching methods, including KMB \cite{zhu2016solving} and LBF \cite{das2019gradient}, when $n$ is set to $32$, $128$, and $512$, respectively.
The matching budget constraints are given by $\rhos=9$ and $\rhot=5$, that is, each task can be assigned to at most $9$ agents and each agent can undertake at most $5$ tasks.
We randomly set $r=10\%$ of the points in $\SM$ as prioritized points.
The cost of matching $s_i$ to $t_{i'}$ follows a uniform distribution from $0$ to $1$.
Given a matching matrix $\TB=[T_{ii'}]$, we evaluate the Proportion of Prioritized Points' Matches out of all matches (PPPM), as well as the Proportion of Satisfied Matching Budgets for Prioritized Points (PSMBPP), defined as follows:
\begin{equation*}
	\resizebox{\linewidth}{!}{$
		\displaystyle
	\begin{aligned}
		\operatorname{PPPM}(\TB)&=\frac{\big|\{(i,i')\mid T_{ii'}\neq0\text{ and }i\text{ is a prioritized point}\}\big|}{\big|\{(i,i')|T_{ii'}\neq0\}\big|},\\
		\operatorname{PSMBPP}(\TB)&=\frac{\big|\{(i,i')\mid T_{ii'}\neq0\text{ and }i\text{ is a prioritized point}\}\big|}{\rhos mr}.
	\end{aligned}
	$}
\end{equation*}
In \name{}, $\aB$ is set according to Eq. (\ref{eq:priority_a}), and $\bB$ is set as a uniform distribution.
It is easy to verify that (\ref{eq:priority_verification}) holds for all $h\in\llbracket \rhos-1\rrbracket$.

\paragraph{Results in prioritized task assignment.}
The experimental results are reported in Table \ref{tab:prioritized_task_assignment}.
The PPPM values of KMB and LBF is close to $10\%$, which implies that they treat all points equally.
\name{} achieves higher PPPM and PSMBPP values than KMB and LBF, which verifies that \name{} can perform prioritized matching.
One can also observe that \name{} always has PSMBPP higher than $\frac{h}{\rhos}$, which is consistent with Theorem \ref{thm:priority}.
\name{} is also orders of magnitude faster than KMB and LBF.

\section*{Conclusion}
In this paper, we proposed a practical OT-based many-to-many matching method based on two components: the $\ell_0$ constraints (matching budget) on each row and column in a transport plan, and the recently-proposed deformed $q$-entropy regularization.
The former sparsifies a matching, while the latter encourages a point to maximally meet the matching budget constraints.
A theoretically guaranteed algorithm was then derived to determine a transport plan effectively and efficiently.
Experimental results on various tasks demonstrate the proposed method successfully extracts many-to-many matchings and achieves good performance.

\bibliographystyle{IEEEtran}
\bibliography{manymany}

\clearpage

%

%


\end{document}